\documentclass{article}

\usepackage{microtype}
\usepackage{graphicx}
\usepackage{subfigure}
\usepackage{booktabs} % for professional tables

\usepackage{enumitem}
\usepackage[normalem]{ulem}

\usepackage{hyperref}

\usepackage[accepted]{icml2024}

\usepackage{amsmath}
\usepackage{amssymb}
\usepackage{mathtools}
\usepackage{amsthm}
\usepackage{pifont}
\usepackage{colortbl}

\usepackage[capitalize,noabbrev]{cleveref}

\theoremstyle{plain}

\theoremstyle{definition}

\theoremstyle{remark}

\usepackage[textsize=tiny]{todonotes}

\definecolor{pptbrown}{RGB}{132,60,12}
\definecolor{pptgreen}{RGB}{56,87,35}
\definecolor{pptred}{RGB}{155,30,20}
\definecolor{pptdy}{RGB}{127,96,0}

\newcommand{\parh}[1]{\noindent\textbf{#1}}

\newcommand{\tool}{\textsc{SupCBM}}

\newcommand{\F}{Fig.}
\renewcommand{\S}{Sec.}
\newcommand{\T}{Table}

\newcommand{\cls}{$\{\text{CLS}\}$}
\newcommand{\cep}{$\{\text{CEP}\}$}
\newcommand{\descr}{$\{\text{DESCR}\}$}

\begin{document}
\twocolumn[
\title{Eliminating Information Leakage in Hard Concept Bottleneck Models with Supervised, Hierarchical Concept Learning}

\author{Ao Sun, Yuanyuan Yuan$^*$, Pingchuan Ma, and Shuai
  Wang$^*$\\
  The Hong Kong University of Science and Technology
  \\
  \tt \{asunac, yyuanaq, pmaab, shuaiw\}@cse.ust.hk 
}

\date{}
\maketitle
]
\footnotetext[1]{$^*$Corresponding authors.}

\icmlsetsymbol{equal}{*}

\icmlkeywords{Machine Learning, ICML}

\vskip 0.3in

\begin{abstract}
    Concept Bottleneck Models (CBMs) aim to deliver interpretable and
    interventionable predictions by bridging features and labels with
    human-understandable concepts. While recent CBMs show promising potential,
    they suffer from information leakage, where unintended information beyond
    the concepts (either when concepts are represented with probabilities
    or binary states) is leaked to the subsequent label prediction. Consequently,
    distinct classes are falsely classified via \textit{indistinguishable} concepts,
    undermining the interpretation and intervention of CBMs.

    This paper alleviates the information leakage issue by introducing label
    supervision in concept prediction and constructing a hierarchical concept
    set. Accordingly, we propose a new paradigm of CBMs, namely \tool, which
    achieves label prediction via predicted concepts and a
    deliberately-designed intervention matrix. \tool\ focuses on concepts that
    are mostly relevant to the predicted label and only distinguishes classes
    when different concepts are presented. 
    Our evaluations show that \tool\ outperforms SOTA CBMs over diverse
    datasets. It also manifests better generality across different backbone
    models. With proper quantification of information leakage in different CBMs,
    we demonstrate that \tool\ significantly reduces the information leakage.

\end{abstract}

\section{Introduction}

Deep neural networks (DNNs) have been a game-changer in the field of
artificial intelligence (AI), exhibiting remarkable performance in
various machine learning tasks such as computer vision, natural language
processing, and speech recognition. Even though these networks have enabled
a strong performance in many tasks, they are often considered as black-box
models since their extracted features are often obscure and hardly interpretable.

To address this issue, concept bottleneck models (CBMs) have recently emerged
with the purpose of delivering high-quality explanations for DNN
predictions~\cite{koh2020concept}. CBMs are a type of DNN that makes predictions
based on human-understandable concepts. CBMs typically have a Concept-Bottleneck
(CB) layer located before the last fully-connected (FC) layer. The CB layer
takes features extracted (by preceding layers) from the input and maps them to a
set of concepts. This effectively aligns the intermediate layers of a DNN with
some pre-defined expert concepts, and the last FC determines the final label
over those concepts. Often, CBMs require first training the CB layer to align
each neuron to a concept that is pre-defined and understandable to humans.

Based on the output concepts of the CB layer, CBMs deliver two key
benefits: \textit{interpretability} and \textit{intervenability}. First,
users can interpret the predicted labels by inspecting the involved
concepts. Second, users can intervene the predicted labels by modifying
which concepts are involved in the prediction.
Technically, CBMs are often categorized as soft CBMs and hard CBMs, depending on
how concepts are represented and contributed to the label predictions. The CB
layer in soft CBMs outputs a probability (i.e., a number between 0 and 1) for
each concept, whereas the hard CB layer outputs a binary state (i.e., 0 or 1) to
indicate if a concept exists in the input. Then, a label predictor (often the
last FC layer) takes the concept probabilities or states as input and predicts
the final label.

Despite the encouraging potential of CBMs in delivering human-understandable
explanations, their interpretability and intervenability are largely undermined
in de facto technical solutions due to \textit{information leakage}, where unintended
information beyond the concepts is exploited by the label predictor.
Specifically, the concept probabilities in soft CBMs may encode class
distribution information, such that the label predictor can classify
distinct class labels based on \textit{indistinguishable} concepts~\cite{havasi2022addressing}.
For instance, the label predictor may leverage the probability differences of
``\texttt{head}'' and ``\texttt{tail}'' to classify dog vs. cat and achieve
high accuracy, despite that these two concepts are insufficient to distinguish
dog and cat.
While hard CBMs were previously believed resilient to information leakage,
a recent study~\cite{mahinpei2021promises} has pointed out that hard CBMs can
exploit unrelated hard concepts to convey class distributions to the
label predictor. For example, \cite{mahinpei2021promises} demonstrate that
hard CBM's performance can be improved by adding \textit{meaningless} hard
concepts.

To faithfully achieve CBM's design objectives, it is urgent to
mitigate information leakage in CBMs. To this end, this paper proposes a
novel paradigm of CBMs, dubbed as \tool. Unlike previous CBMs that treat
concept prediction and label prediction as two separate tasks, we fuse them
into a unified form by additionally supervising the concept prediction with
class labels. Specifically, we do not implement the label predictor in \tool.
Rather, we maintain an intervention matrix to determine which concepts should
be involved for a given class label. The intervention matrix is formed when
constructing the concept set; it is implemented as a sparse binary matrix of
shape \#concepts $\times$ \#classes, where the $(i, j)$-th entry indicates
whether the $i$-th concept should be leveraged to recognize the $j$-th class. 

When training \tool, the CB layer is forced to only predict concepts that are
relevant to the ground truth label, which is jointly decided by the intervention
matrix and a novel concept pooling layer (which further selects the most important
concepts for each input; see \S~\ref{subsec:supervision}). Then, to obtain the
final predicted label, we multiply the CB layer's output, a
\#concepts-dimensional vector indicating the predicted concepts, with the
intervention matrix. This computation is equivalent to summing up the involved
concepts' probabilities for each label and treating the summed probability as
the label's prediction confidence. This label prediction procedure only
distinguishes classes if different concepts are presented. As demonstrated in
\S~\ref{subsec:supervision} and \S~\ref{subsec:eval-leakage}, \tool\ can
significantly alleviate the information leakage. Moreover, similar to previous
post-hoc CBMs like \citet{oikarinen2023label}, \tool\ also achieves post-hoc
CBMs by only training a light-weight CB layer (without training any label
predictor) for any pre-trained feature-based model.

Besides reformulating the CBM technical pipeline, we augment CBM
interpretability and intervenability from the concept aspect. We argue that the
concept set should prioritize perceptual concepts that can be directly perceived
by humans \textit{without} further reasoning, e.g., ```\texttt{tail}'' instead
of ``\texttt{animal}''. Also, to fully use the rich semantics of these
perceptual concepts (which are mostly nouns), we assign each of them multiple
descriptive concepts (i.e., adjectives) and build a two-level hierarchical
concept set. With the novelly formed concept set, \tool\ outperforms all SOTA
CBMs and even reaches the vanilla feature-based model (w/o converting features
into concepts) when evaluated using diverse datasets and backbones (see
\S~\ref{sec:evaluation}).
Overall, this paper makes the following contributions:

\begin{itemize}[leftmargin=*, topsep=0pt, itemsep=0pt, parsep=0pt]
    \item We propose a new paradigm of CBMs, \tool, where the concept
    prediction is directly supervised by the class label and the
    follow-up label predictor is not required. \tool\ significantly
    reduces information leakage in previous CBMs and also delivers 
    post-hoc CBMs.
    
    \item We advocate that the concept set should be primarily built
    with perceptual concepts that can be directly perceived by humans
    without further reasoning. To utilize rich semantics in perceptual
    concepts, we build a two-level hierarchical concept set by assigning
    each perceptual concept with multiple descriptive concepts.

    \item Evaluations show that \tool\ eliminates information leakage and offers
    highly interpretable and intervenable predictions. \tool\ also outperforms
    previous SOTA CBMs and achieves comparable performance to the vanilla,
    feature-based models.
\end{itemize}
\section{Background and Motivation}
\label{sec:background}

This section presents background of CBMs and reviews various implementations of
CBMs in prior works.

Without losing the generality, we take image classification as an example in
this section.

\subsection{Concept Bottleneck Models}
\label{subsec:cbm}

\parh{Dataset Construction.}~Given a conventional dataset
$D = \{ x^{(i)}, y^{(i)}\}_{i=1}^{N}$ consisting of $N$ pairs of input image
$x^{(i)} \in \mathbb{R}^{d}$ and its ground truth label $y^{(i)} \in \mathbb{N}$,
CBMs require augmenting the dataset $D$ with annotated concepts for
all images. Specifically, for a set of pre-defined concepts $C$, each image
$x^{(i)}$ is annotated using $c^{(i)} \in \{0, 1\}^{|C|}$. If the $j$-th element
of $c^{(i)}$ is 1, it indicates that the $j$-th concept in $C$ is presented in
$x^{(i)}$, and vice versa.

The concept set $C$ is often manually defined by humans and the annotations of
$c^{(i)}$ require extensive human efforts, limiting the applicability of CBMs.
Recent works~\cite{oikarinen2023label,yang2023language} have employed LLMs to
automatically generate concepts and automate the annotation process with
similarity scores from multimodal alignment models (e.g., CLIP), bringing richer
and more expressive concepts sets to CBMs. However, we note that their concepts
are often obscure, which impedes interpreting and intervening CBMs. This paper
alleviates this issue by focusing on perceptual concepts and their
descriptions.

\parh{Soft and Hard CBMs.}~CBMs divide the label prediction $y = \arg\max f(x)$
into two main steps: concept prediction $c = g(x)$, which is achieved by
appending the penultimate in $f$ with the CB layer, and label prediction
$y =\arg\max h(c)$.
Accordingly, CBMs require additionally training the concept predictor, which
can be conducted prior to or jointly with training the label predictor.
The training of concept predictor is often formulated as multiple
concurrent binary classification tasks, with each one for one concept.

During CBM's inference, the concept predictor outputs a probability for each
concept. If the follow-up label prediction directly takes these probabilities,
the CBM is categorized as \textit{soft CBM}. In contrast, if the CBM first
converts the probabilities to binary states (i.e., 0 or 1) and then feeds
them to the label predictor, the CBM implements \textit{hard CBM}. 

\parh{Side-Channel CBMs.}~Soft CBMs are more frequently used due to their
plausible performance (comparable to feature-based models). However, information
leakage~\cite{havasi2022addressing}, where concept probabilities encode
unintended information beyond the concepts themselves, undermines the
interpretability and intervenability of soft CBMs. Previous works
therefore leverage hard CBMs to alleviate this issue. Since hard CBMs
usually exhibit limited performance. Previous works propose side-channel
CBMs~\cite{havasi2022addressing} by adding ``side channels''
(residual connections) to hard CBMs, which link the concept
predictor and the label predictor, and passes additional information to the
label predictor. While the side channel can improve hard CBM's performance,
the interpretability and intervenability are simultaneously
sacrificed~\cite{havasi2022addressing}.

Worse, as pointed out by~\citet{mahinpei2021promises}, the CB layer in hard
CBMs can also leak unintended information to the label predictor via
irrelevant (hard) concepts, rendering the high demand of a new CBM paradigm
that is resilient to information leakage. As will be demonstrated in
\S~\ref{subsec:supervision} and \S~\ref{subsec:eval-leakage}, our novel CBM, \tool, can significantly reduce
the information leakage.

\parh{Post-hoc CBMs.}~When converting standard feature-based models into CBMs, the
common practice is to tune the whole model. Nevertheless, this incurs a high
training cost due to the large number of parameters in the concept predictor.
Post-hoc CBM~\cite{yuksekgonul2022post} is therefore proposed to efficiently
convert pre-trained backbone models into CBMs.

In particular, post-hoc CBM first builds a concept bank and forms
a concept subspace with the embeddings of these concepts. Then, the embedding
generated by the backbone model is first projected onto the concept subspace,
and then the predicted concepts (e.g., stripe, horse) are fed into the label
predictor (e.g., zebra). Post-hoc CBMs retain the weights in the original model,
and only train the label predictor. Therefore, they are more efficient and 
applicable to large-scale models.

\tool\ achieves post-hoc CBM by only training a light-weight one-layer CB layer
for each backbone model. Thus, \tool's label prediction is conducted by
multiplying the concept prediction with an intervention matrix, which requires
no training (see \S~\ref{subsec:supervision}). Moreover, while previous CBMs and
\tool\ are both deemed post-hoc, evaluations in \S~\ref{subsec:eval-generality}
shows that previous post-hoc CBMs manifest lower generality than \tool\ when
applied to different backbone models.

\section{\tool's Design}
\label{sec:design}

\begin{figure*}[!ht]
    \centering
    \includegraphics[width=0.92\linewidth]{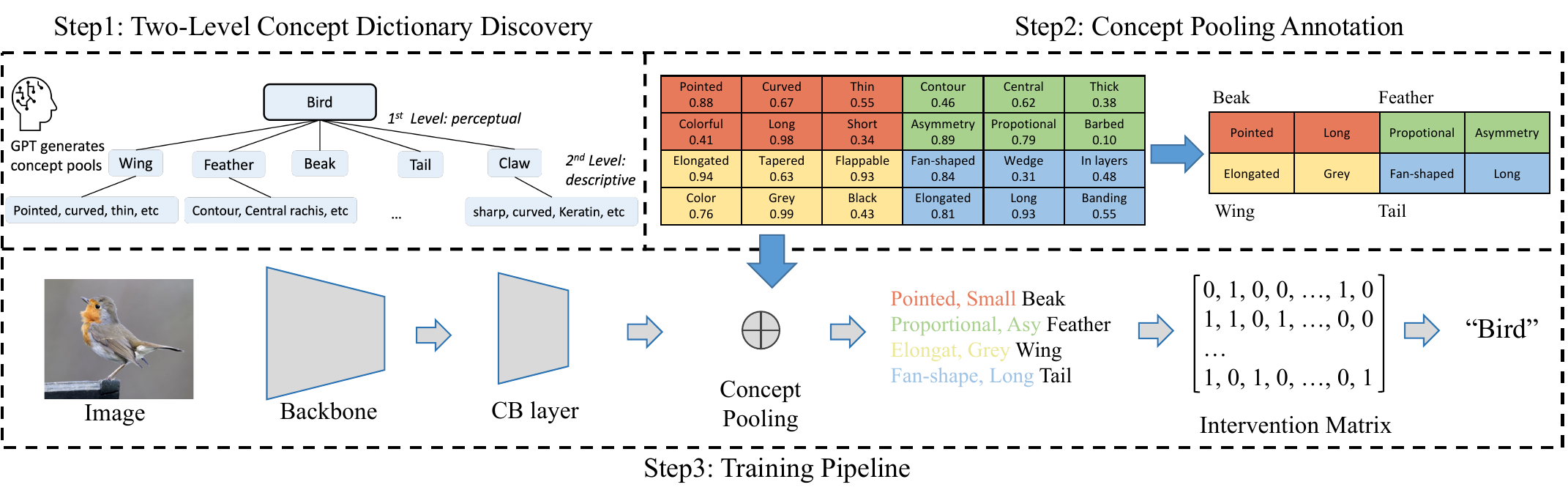}
    \vspace{-10pt}
    \caption{Workflow of \tool. The training of \tool\ consists
    three mains stages.}
    \label{fig:workflow}
    \vspace{-5pt}
\end{figure*}

\parh{Technical Pipeline.}~\F~\ref{fig:workflow} illustrates the workflow of
\tool, which consists of three main stages. In the concept construction stage, we
leverage GPT (version 4.0) to generate a comprehensive set of concepts and organize them in a
two-level hierarchical manner. Meanwhile, we also maintain an intervention
matrix to record each concept's relevant concepts. Then, during the training
stage, we train the CB layer to map the feature extracted by the backbone model
into our prepared concepts. To reduce the concept leakage, the training is
supervised by the ground truth label and a concept pooling layer. Finally, when
predicting the class labels, we multiply the CB layer's concept prediction with
our intervention matrix; the final label is obtained by applying $\arg\max$ on
the multiplication result.

\subsection{Concept Set Construction}
\label{subsec:construction}

\parh{Perceptually Perceivable Concepts.}~One key observation made in this
paper, is that some concepts in previous works are often too holistic to be
easily perceivable by humans and require manual reasoning. For instance, the
concept ``\texttt{animal}'' is involved in classifying cats. Although animal is
a genuine super-class of cat, it cannot be perceptually observed; prior
knowledge is required to reason that cat is an animal. Also, such concepts
impede the intervention of CBMs: it is obscure to add or remove the concept
``\texttt{animal}'' for the follow-up label prediction. Thus, we advocate that
\textit{the concept set should be formed with perceptual concepts that can be directly
observed by humans without reasoning.}

\parh{Hierarchical Concepts.}~Since CBMs are designed to only describe whether
certain concepts are involved in the label prediction, the rich semantics of
concepts often cannot be faithfully reflected from CBMs. Considering the case of
classifying different breeds of fish in CIFAR100, while the concept
``\texttt{tail}'' can be leveraged for predicting labels, the information it
contributes is far beyond than simply indicating its involvement. For example, a
``\texttt{fan tail}'' is critical to identify \texttt{goldfish}, whereas the
``\texttt{veil tail}'' is vital to identify \texttt{betta fish}. 

Thus, we form a two-level hierarchical concept set, where the first level
contains nounal perceptual concepts and primarily flags if a concept is
involved in the label prediction, whereas the second level includes adjectival
concepts for each first-level concept and describes which semantics of the
visual concepts contribute to the label prediction.

\parh{Prompt Design.}~We use GPT to automatically build the concept set. Since
we require perceptually perceivable concepts to enhance CBM interpretability and
intervenability, we first query the GPT model with the following prompt:
\begin{quote}
    ``\textit{To identify \cls\ visually, please list the most important $\{p\}$
    \textbf{visual} parts which a \cls\ has.}''
\end{quote}
\vspace{-5pt}

Here, the token \cls\ can be replaced by any classes the original feature-based
model can predict. The GPT returns a set of nounal visual parts in \cls\ as our
first-level concept. Then, for each concept \cep\ of a class \cls, we query the
GPT model with the following prompt to uncover semantics associated in different
concepts.

\vspace{-5pt}
\begin{quote}
    ``\textit{To visually identify \cls, please describe the $\{q\}$ most common
    characteristics of \cls's \cep\ from the three dimensions of
    shape, color, or size.}''
\end{quote}
\vspace{-5pt}

With the above prompt, the GPT model can return to us a set of adjectival
concepts as our second-level concepts that describe the semantics of the
first-level nounal concepts. We limit the adjectival concepts to follow
the three dimensions of shape, color, and size, thereby reducing ambiguity.
We also remove any concept that is longer than 40 characters to keep the
descriptions simple.
This way, we obtain a two-level concept set containing concepts represented in a
$\langle \cep, \descr\ \rangle$ form, as illustrated in \F~\ref{fig:workflow},
e.g., $\langle Pointed, Beak \rangle$ constitutes one $\langle \cep, \descr\
\rangle$ tuple for the input \texttt{bird} image.
Accordingly, we also
know each $\langle \cep, \descr\ \rangle$'s relevant class \cls.

\parh{Extendability.}~Our concept set is highly extendable. To convert
any backbone models whose supported classes are not included in our
concept set, users can simply replace the \cls\ in the first prompt
and query GPT models to obtain the corresponding concepts, and accordingly
update the \cls\ and \cep\ in the second prompt to augment the concept
set with new $\langle \cep, \descr\ \rangle$ pairs.

\subsection{Training Supervision}
\label{subsec:supervision}

\parh{Intervention Matrix.}~We maintain an intervention matrix to
represent which concepts should be involved in predicting a class label.
The intervention matrix is a binary matrix of shape \#concepts $\times$ \#classes,
as shown below,
\begin{equation}
\small
\begin{bmatrix}
    \mathcal{I}_{1, 1} & \mathcal{I}_{1, 2} & \dots & \mathcal{I}_{1, \text{\#classes}} \\
    \mathcal{I}_{2, 1} & \mathcal{I}_{2, 2} & \dots & \mathcal{I}_{2, \text{\#classes}} \\
    \vdots & \vdots & \vdots & \vdots \\
    \mathcal{I}_{\text{\#concepts}, 1} & \mathcal{I}_{\text{\#concepts}, 2} & \dots & \mathcal{I}_{\text{\#concepts}, \text{\#classes}} \\
\end{bmatrix}
\end{equation}

\noindent where $\mathcal{I}_{i,j} \in \{0, 1\}$ and $\mathcal{I}_{i,j} = 1$
indicates that the $i$-th concept should be involved in predicting the
$j$-th class; otherwise, $\mathcal{I}_{i,j} = 0$. This is
obtained after querying the GPT with the second prompt mentioned in
\S~\ref{subsec:construction}.

\parh{Label-Aware Concept Annotation.}~A fundamental difference between
\tool\ and previous CBMs is the supervision of class label when training
the CB layer. As illustrated in \F~\ref{fig:workflow}, for each training
image $x$ whose ground truth label is $y$, we annotate it only with
concepts that are involved in predicting $y$ (as indicated by the
intervention matrix). This annotation, to some extent, can supervise
the CB layer to ``fuse'' the concept prediction and label prediction,
which can improve the performance of CBMs and simultaneously reduce
the information leakage; see evaluations in \S~\ref{sec:evaluation}.

\parh{Concept Pooling.}~Since~\citet{mahinpei2021promises} point out that
adding irrelevant concepts (even in a hard form) can enlarge information
leakage, we further implement a selective strategy as shown in \F~\ref{fig:workflow}.
Similar to the max pooling mechanism in conventional computer vision which
selects most important features, we choose to annotate $x$ with those most
important concepts. We name our selection procedure as concept pooling
as it is equivalent to applying an 1-dimensional max pooling of kernel
size $q$ and stride $q$ on all second-level concepts.
Specifically, we first compute $x$'s similarities with all concepts
$\langle \cep, \descr\ \rangle$ that are involved in predicting $y$.
To do so, we measure the cosine similarity between the CLIP embeddings
of $x$ and $\langle \cep, \descr\ \rangle$.
Then, for concepts $\langle \cep, \descr\ \rangle$ sharing the same \cep,
we choose those having the top-$k$ similarity as the ground truth concepts.
This way, we have total $p \times k$ concept annotations for each input image.

\parh{Training Objectives.}~Unlike previous post-hoc CBMs that directly train the
follow-up label predictor with concept similarities, we do not train a label
predictor. Instead, we train the CB layer (without training the backbone model)
since label supervision is unavailable for test images. Specifically, for each
input image, we set the selected $p \times k$ concepts as hard label and form $p
\times k$ binary classification tasks following the conventional training
procedure. These binary classification tasks are optimized with binary
cross-entropy loss $\ell_{BCE}(c, GT_c)$, where $c$ is the concept probabilities
predicted by the CB layer and $GT_c$ is the ground truth concepts. Both $c$ and
$GT_c$ are  $pq$-dimension vectors.

The predicted label is decided as $\arg\max c * \mathcal{I}$, where $*$
denotes matrix multiplication. That is,
the $j$-th class's prediction confidence is computed as
\begin{equation}
    l_j = \sum_{i=1}^{pq} c_i \cdot \mathcal{I}_{i,j}
\end{equation}

\noindent Since $\mathcal{I}_{i,j}$ indicates whether the $i$-th concept
should be involved in predicting the $j$-th class, the label $j$'s confidence
$l_j$ equals the sum of probabilities of those involved concepts.
Consistent with conventional CBM training, we have a cross-entropy loss
$\ell_{CE}(l, GT_l)$, where $l$ is a \#classes-dimensional vector and
$GT_l$ is an integer indicating the ground truth class label. It's worth
noting that the intervention matrix $\mathcal{I}$ is always fixed,
and $\ell_{CE}(l, GT_l)$ is only leveraged to optimize the CB layer.
Therefore, the overall training objective is to minimize:
\begin{equation}
    \alpha \cdot \ell_{BCE}(c, GT_c) + (1 - \alpha) \cdot \ell_{CE}(c * \mathcal{I}, GT_l)
\end{equation}

\noindent where $\alpha$ is a hyper-parameter.

\parh{Eliminating Information Leakage.}~\tool\ eliminates information leakage
from two aspects. First, for hard information leakage where irrelevant concepts
may leak unintended information for label prediction, we inherently mitigate it
during the concept annotation. As mentioned above, when annotating concepts, we
only focus on those concepts that are involved in predicting the label. Second,
we further select those most important descriptive concepts \descr\ via our
concept pooling mechanism, thus trimming irrelevant concepts.
With these two steps, irrelevant concepts are rarely kept in label predictions,
reducing hard information leakage noted in~\citet{mahinpei2021promises}.

Moreover, for soft information leakage where the concept probabilities encode
distribution of class information, we justify how our intervention matrix
can eliminate such leakage. Considering two classes $a$ and $b$ whose
involved concepts (according to $\mathcal{I}$) constitute sets $C_a$ and
$C_b$, respectively. We define  $S = C_a \cap C_b$, $A = C_a \setminus S$,
and $B = C_b \setminus S$. That is, $S$ consists of concepts shared by both
classes, and $A$ and $B$ denote the sets of concepts that are only involved in
predicting $a$ and $b$, respectively.

\ding{172}~If $S = \emptyset$, i.e., the two classes do not share any concepts, it
should be clear that classifying $a$ and $b$ only rely on their disjoint
concepts $A$ and $B$, and \tool\ finds concepts that uniquely exist in classes
$a$ and $b$. Note that our concept pooling mentioned above can help reduce
the size of $S$.

And in case $S \neq \emptyset$, i.e., the two classes share some concepts, we
analyze the following three cases.

\ding{173}~If $A = \emptyset$, i.e., the class $a$ does not have any unique concept,
when the input's label is $a$, the label prediction confidences of
class $a$ and $b$ should be identical. Thus, the label prediction
(i.e., $c * \mathcal{I}$ in \tool) cannot distinguish $a$ and $b$,
and no information beyond the concept is exploited to falsely conduct
classification. Similar conclusions can be drawn if $B = \emptyset$
or $A = B = \emptyset$. 

\ding{174}~If $A \neq \emptyset$ and $B \neq \emptyset$, i.e., both classes have
their unique concepts. Then given an input, the difference between its
two label prediction confidences (w.r.t. class $a$ and $b$) is only
induced by the concepts belonging to $A \cup B$. Accordingly, no information
beyond $a$ and $b$'s unique concepts is leveraged to classify them.

From the above three cases, we can safely conclude that \tool\ eliminates
soft information leakage.

\parh{Post-hoc Concept Mapper.}~It's worth noting that \tool\ also achieves
post-hoc CBMs. Similar to existing post-hoc CBMs, to convert a new
backbone model, \tool\ only requires training a light-weight layer. Differently,
previous works train the label predictor using their pre-built concept set,
while \tool\ only trains an one-layer CB layer without training any label
predictor.

\section{Implementation}
\label{sec:implementation}

\parh{Concept Set Construction.}~When querying GPT (as introduced in
\S~\ref{subsec:construction}), we set $p=5$ for the first prompt and
$q=6$ for the second prompt. That is, for each class, we obtain $5$
perceptual concepts with each one having $6$ descriptive concepts.
When annotating the concepts using our concept pooling, we set $k=2$.
That is, each training input is annotated with total ($5$ perceptual
$\times$ $2$ descriptive) concepts.

\parh{Hyperparameters.}~For all experiments, we use the Adam optimizer to
optimize the loss function, with $\alpha=0.7$. In accordance with the
previous works~\cite{yuksekgonul2022post, yang2023language}, we tune 
other hyperparameters on the development set.

\section{Evaluations}
\label{sec:evaluation}

We conduct performance evaluation in \S~\ref{subsec:eval-performance}, and
assess information leakage in \S~\ref{subsec:eval-leakage}. We also benchmark
the generality of \tool\ in \S~\ref{subsec:eval-generality}, and
\S~\ref{subsec:case} presents case studies of \tool's concepts and
interpretations.

\subsection{Setup}
\label{subsec:setup}

Following previous
works~\cite{yuksekgonul2022post,oikarinen2023label,yang2023language}, we
consider the following representative datasets to evaluate the performance of
different models. These datasets are representative and cover three major
computer vision domains including general classification (CIFAR10, CIFAR100),
specialized classification (CUB-Bird), and medical image analysis (HAM10000).

\textbf{CIFAR-10}~\cite{krizhevsky2009learning} consists of 32$\times$32 RGB-color images of
10 classes and each class has 6,000 images.

\textbf{CIFAR-100}~\cite{krizhevsky2009learning} consists of 32$\times$32 RGB-color images of
10 classes and each class has 600 images.

\textbf{CUB-Bird}~\cite{wah2011caltech} consists of 11,788 RGB-color images of 200 bird
species. 

\textbf{HAM10000}~\cite{tschandl2018ham10000} consists of 10,015 dermatoscopic images of
pigmented skin lesions.

For CIFAR10 and CIFAR100, all CBMs use the same backbone CLIP-RN50 following
in~\citet{yuksekgonul2022post,oikarinen2023label,yang2023language}.
For CUB-Bird, following~\citet{yang2023language}, we use CLIP-ViT14 as the
backbone model to obtain a high performance. Consistent
with~\citet{oikarinen2023label}, we also ignore grayscale images in CUB, and only
consider the RGB format, which results in 5,990 training images and 5,790 test
images. For HAM10000, we use the standard HAM-pretrained
Inception~\cite{daneshjou2021disparities} as the backbone.

\subsection{Performance Comparison}
\label{subsec:eval-performance}

We compare our method with the following SOTA CBMs.

\noindent \textbf{P-CBM} \cite{yuksekgonul2022post} is the first post-hoc CBM that uses the CLIP
embeddings to align images and concepts. P-CBM first projects an image embedding
onto the concept subspace and then computes its similarities with different concepts.
These similarities are adopted for label prediction. The label predictor in P-CBM
is connected with image embeddings via a residual connection.

\noindent \textbf{Label-free CBM} \cite{oikarinen2023label} is mostly similar to
P-CBM, but computes similarity between images and concepts using the dot product
of their embeddings. Label-free CBM is the first work that employs GPT models
to generate textual concepts.

\noindent \textbf{LaBo} \cite{yang2023language} builds semantic vectors with a
large set of attributes from LLMs. It uses GPT-3 to produce factual sentences
about categories to form candidate concepts. It then employs a so-called
submodular utility to effectively explore potential bottlenecks that facilitate
the identification of distinctive information.

\noindent \textbf{Yan} \cite{yan2023learning} uses concise and descriptive
attributes extracted from LLMs. Specifically, it uses a learning-to-search
method to extract a descriptive subset of attributes from LLMs by pruning the
large attribute set.

Besides the above CBMs, existing works also evaluate a setting (referred to as
\textbf{Feat}), which directly uses features extracted by the backbone model for
the subsequent label prediction. In short, \textbf{Feat} can be deemed to offer
the ``upper bound'' performance of CBMs.

\begin{table}[!htbp]
\vspace{-10pt}
\caption{Performance comparison with SOTA CBMs and the vanilla setting
\textbf{Feat}. We mark the best performance in \colorbox{blue!25}{blue} and the
second best in \textbf{bold}. ``N/A'' indicates that they do not provide the
related concept-generating implementation.}
\label{tab:performance}
\vspace{-10pt}
\begin{center}
\begin{small}
\begin{sc}
	\resizebox{1\linewidth}{!}{
\begin{tabular}{lcccc}
\toprule
Model & CUB-Bird & CIFAR10 & CIFAR100 & HAM10000 \\
\midrule
Feat   & \colorbox{blue!25}{86.41}& \textbf{88.80} & \textbf{70.10} &  \colorbox{blue!25}{84.07} \\
\midrule
P-CBM & 78.31  & 84.50& 56.60 & 70.69\\
Label-Free    & 78.88&86.40& 65.13 & 72.79 \\
Labo     & 83.82 & 87.90&69.10 & 82.03\\
Yan      & 81.67 & 79.31 & 66.66 & N/A \\
\tool\    & \textbf{85.98} &  \colorbox{blue!25}{89.22}&  \colorbox{blue!25}{70.11} & \textbf{83.28}        \\
\bottomrule
\end{tabular}
    }
\end{sc}
\end{small}
\end{center}
\vskip -0.1in
\end{table}

\T~\ref{tab:performance} reports the performance of all SOTA CBMs, the vanilla
feature representation (\textbf{Feat}), and \tool, on the four datasets. We see
that \tool\ constantly outperforms all SOTA CBMs across all datasets. Moreover,
\tool\ exhibits comparable performance with \textbf{Feat}, and even outperforms
it on the CIFAR10 and CIFAR100 datasets. We interpret the findings as highly
encouraging, demonstrating the superiority of our new CBM paradigm.

\subsection{Information Leakage}
\label{subsec:eval-leakage}

\parh{Setup.}~\S~\ref{subsec:eval-performance} shows that LaBo has the best
performance among all our competitors. Therefore, this section assesses the
information leakage in LaBo and our method. We use CIFAR100, given the large
number of distinct classes this dataset has. We also consider the Flower
dataset~\cite{nilsback2008automated}; it is a phytology knowledge specific
dataset with 102 visually close semantic classes, making it handy to quantify
the information leakage problem.

\parh{Metrics \& Intuitions.}~Since a quantitative metric of information leakage
is still lacking, we propose a new metric to assess the information leakage in
CBMs. The information leakage problem is induced by the insufficient concept
set, such that the label prediction has to exploit unintended information to
fulfill the training objective (e.g., achieving high
accuracy)~\cite{mahinpei2021promises}. In that sense, if a CBM suffers from
information leakage, removing concepts that contribute most to the label
prediction should notably undermine the CBM performance.
Following this intuition, for each evaluated CBM, we start by training it with
the full concept set. Then, during the inference, we rank concepts based on
their importance and gradually remove top-ranked concepts. We expect that the
performance of CBMs which manifest better resilience to information leakage
should drop more quickly when more concepts are removed. Here, we use our
constructed concept set for a fair comparison. 

\parh{Baselines.}~We set the baseline as a dummy linear model which does not
have concept alignment as CBMs (i.e., the ``concepts'' are purely random) and
should have the most severe information leakage issue. Also, as an ablation, we
replace the intervention matrix in \tool\ with a learned fully-connected layer.
We denote the ablated version of \tool\ as $\text{\tool}_{FC}$.

\begin{figure}[!ht]
    \centering
    \vspace{-5pt}
    \includegraphics[width=1.00\linewidth]{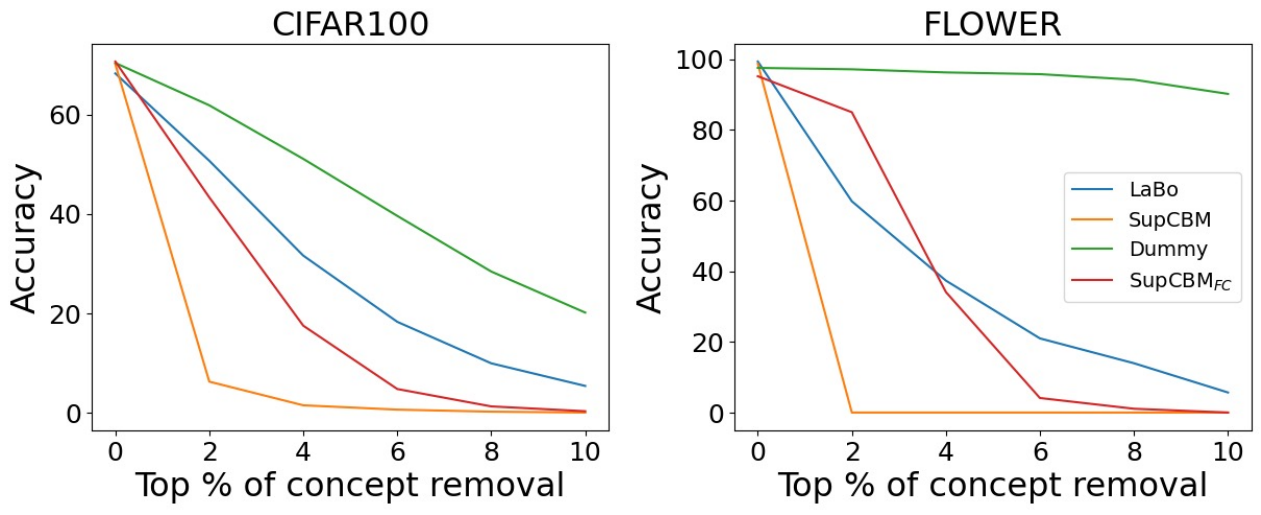}
    \vspace{-20pt}
    \caption{Information leakage evaluation. If a CBM manifests
    higher resilience to information leakage, its performance
    should drop more quickly when more concepts are removed.}
    \label{fig:leak}
    % \vspace{-5pt}
\end{figure}

\parh{Results.}~As shown in \F~\ref{fig:leak}, we can see that the dummy model manifests
the lowest resilience to information leakage, as its performance drops slowly
when more concepts are removed. Specifically, even when top 10\% important
concepts are removed, the dummy model still has $\sim$20\% and $\sim$90\%
accuracy for the CIFAR100 and the Flower datasets, respectively. 
\F~\ref{fig:leak} also shows that \tool's performance drops the fastest when
more concepts are removed in both two datasets, indicating its highest
resilience towards information leakage. In addition, when cross-comparing
$\text{\tool}_{FC}$ with LaBo, $\text{\tool}_{FC}$ is lowered to random guess
more quickly than LaBo. We interpret that the gap is due to our label-supervised
concept prediction, which rules out irrelevant concepts and thus improves the
resilience to information leakage~\cite{mahinpei2021promises}. Moreover, when
cross-comparing \tool\ with $\text{\tool}_{FC}$, it is evident that the label
prediction conducted via the intervention matrix also improves the resilience
to information leakage, which has been justified in \S~\ref{subsec:supervision}.

\subsection{Generality}
\label{subsec:eval-generality}

While recent post-hoc CBMs achieve promising performance, they all take the text
encoder of CLIP as the backbone to label concepts (i.e., obtaining the text
embeddings of concepts); however, the goal of CBMs is to understand the
pair-wised visual backbone model (i.e. the corresponding image encoder). So, it
is unclear whether the concept knowledge learned by the text-encoder will be
fully aligned or unbiased with the pair-wised image encoder. Therefore, we
conduct a generality evaluation here to benchmark these CBMs from the
``post-hoc'' perspective. Specifically, we incorporate three types of pair-wised
text-vision backbone models, including RN-50, ViTB-32, ViTL-14, into these CBMs
and evaluate their performance.

\begin{figure}[!ht]
    \centering
    % \vspace{-5pt}
    \includegraphics[width=1.00\linewidth]{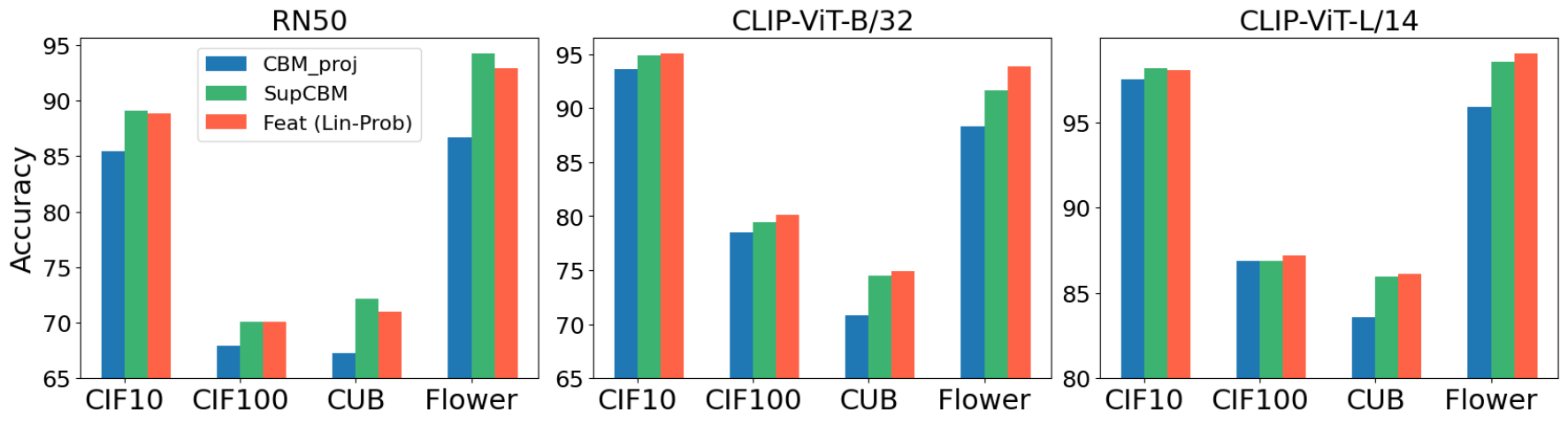}
    \vspace{-20pt}
    \caption{Generality evaluation.}
    \label{fig:general}
    \vspace{-5pt}
\end{figure}

\textbf{Baselines.}~We consider two baselines to compare with \tool:
\textbf{Feat} and \textbf{CBM-proj}. As aforementioned, \textbf{Feat} is the
vanilla feature representation, which offers the ``upper bound'' performance
toward targeted vision backbones. \textbf{CBM-proj} is a direct projection of
the CLIP text-image cosine similarity score onto the label space,
which deems the most straightforward way to validate the bias from the CLIP
Text-encoder for visual-model prediction and subsumes prior
post-hoc CBMs. For a fair comparison, we use the same concept
set in \textbf{CBM-proj} as in our method.

Results are shown in \F~\ref{fig:general}; it is seen that \tool\ constantly
outperforms all competitors on all datasets with different backbone models.
Also, these competitors' performance may change largely
with backbone models, whereas \tool's performance is stable when using different
backbones, indicating its better generality.

\begin{figure*}[!ht]
    \centering
    % \vspace{-5pt}
    \includegraphics[width=0.95\linewidth]{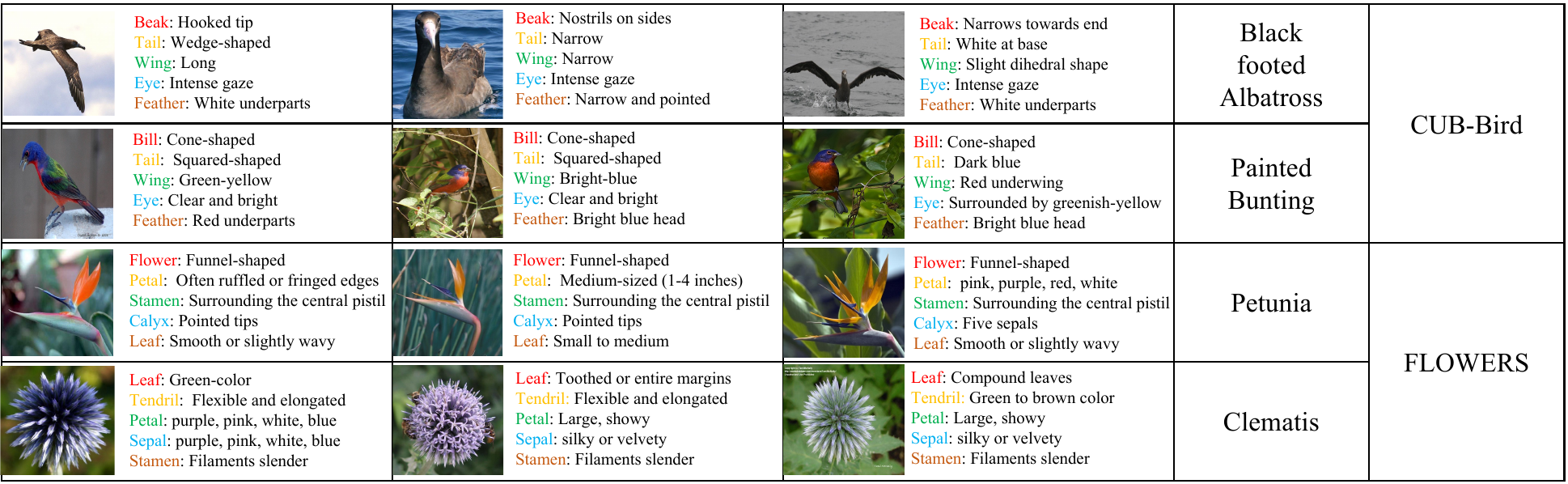}
    \vspace{-5pt}
    \caption{Case study of \tool\ on CUB-Bird and FLOWER datasets.}
    \label{fig:demo}
    \vspace{-5pt}
\end{figure*}

\subsection{Case Studies}
\label{subsec:case}

This section presents cases of \tool's predicted concepts and interpretations
of different predicted classes. We use the CUB-Bird and Flower datasets since they
are finer-grained than other datasets.

Empowered our concept pooling technique, \tool\ can deliver
precise intra-class concept identification.

As shown in \F~\ref{fig:demo}, for each image, we first present the
perceptual concepts (i.e., the first-level concepts) identified by \tool.
Then, for each perceptual concept, we show its most important
(whose probability is the highest) descriptive concept.

Considering the first row of \F~\ref{fig:demo}, where three different
black-footed albatrosses have visually different wings due to their
flying postures, \tool\ can accurately distinguish them by recognizing
the distinct semantics of their wings.
Specifically, the third albatross differs from the other two
with spreading wings, \tool\ can identify such differences by
predicting the ``\texttt{Slight dihedral shape}'' wing.
Also, the second albatross folds its wings, which is faithfully captured
by \tool\ and reflected from the adjectival concept
``\texttt{narrow}'' in both perceptual concepts
``\texttt{tail}'' and ``\texttt{wing}''.

Regarding the Flower dataset shown in the last two rows of \F~\ref{fig:demo},
we find that unlike the CUB dataset where birds have diverse motions, the
concepts (both perceptual and descriptive) are mostly similar for flows
from the same class. However, \tool\ can still capture the subtle 
differences. Considering the Petunia images (the third row) shown in
\F~\ref{fig:demo}, where three images are highly similar in terms of the
color and shape, \tool\ can precisely recognize ``\texttt{five sepals}'' calyx
in the third image, and identify the ``\texttt{Pointed tips}'' calyx in other
two images. Overall, these cases demonstrate the effectiveness of \tool\ in
capturing diverse concepts and rich semantics, and interpreting its
predictions.
\section{Related Works}
\label{sec:related}

Further to \S~\ref{subsec:cbm}, we review other relevant works below.

\parh{Model Interpretation.}~Model interpretation aims to explain the model's
predictions in a human-understandable forms, divided into pixel-based and
concept-based approaches. Pixel-based techniques highlight the input pixels that
contribute most to the model's prediction, including saliency
maps~\cite{simonyan2013deep}, Grad-CAM~\cite{zhou2016learning},
DeepLIFT~\cite{shrikumar2017learning}, GradSHAP~\cite{selvaraju2017grad},
LIME~\cite{ribeiro2016should}, SHAP~\cite{lundberg2017unified}, and
DeepSHAP~\cite{lundberg2017unified}. The interpretation is accomplished by
computing each pixel's (or super-pixel's) contribution to the model prediction.
Nevertheless, the main issue is that such pixel-based interpretation can be hard
to make sense of by humans. Recently, some efforts have been made to improve the
understandability. For instance, EAC~\cite{sun2023explain} delivers
concept-based interpretations by using a segmentation model to generate concept
masks. Similar to CBMs, EAC provides understandable and abstract concepts to
facilitate interpretation of model predictions. However, EAC relies on
approximation of the original model, and thus may fail to faithfully represent
the original model's behavior.

\parh{Model Editing and Repairing.}~Previous model editing/repairing works
primarily focus on generative models (e.g., GANs). By identifying interpretable
direction in the input latent space of generative models, they can achieve
controllable generation of
images~\cite{harkonen2020ganspace,shen2020interpreting}. However, such
techniques do not apply discriminative models like classifiers. Some works in
the software engineering community have proposed to repair DNNs by modifying
their weights or structures. For example, CARE~\cite{sun2022causality} uses
causal reasoning to identify buggy neurons in DNNs and repair them and
NNrepair~\cite{usman2021nn} and provable polytope
repair~\cite{sotoudeh2021provable} use formal methods to repair DNNs. However,
these techniques are orthogonal to CBMs, as they focus on repairing DNNs'
internal structures, while CBMs aim to repair the model's predictions by
modifying the input concepts. Moreover, these techniques are usually
computationally expensive, while CBMs are lightweight as they only modify the
input to the label predictor (i.e., the existence of certain concepts).
\section{Conclusion}
\label{sec:conclusion}

This paper has proposed \tool, a novel approach to improve the interpretability
of CBMs with supervised concept prediction, hierarchical, descriptive  concepts,
and a post-hoc design. Evaluations show that \tool\ can effectively outperform
existing CBMs to offer high interpretability and performance (reaching the
vanilla, feature-based models) across various settings.

\bibliographystyle{icml2024}
\bibliography{main}

%%%%%%%%%%%%%%%%%%%%%%%%%%%%%%%%%%%%%%%%%%%%%%%%%%%%%%%%%%%%%%%%%%%%%%%%%%%%%%%
%%%%%%%%%%%%%%%%%%%%%%%%%%%%%%%%%%%%%%%%%%%%%%%%%%%%%%%%%%%%%%%%%%%%%%%%%%%%%%%
% APPENDIX
%%%%%%%%%%%%%%%%%%%%%%%%%%%%%%%%%%%%%%%%%%%%%%%%%%%%%%%%%%%%%%%%%%%%%%%%%%%%%%%
%%%%%%%%%%%%%%%%%%%%%%%%%%%%%%%%%%%%%%%%%%%%%%%%%%%%%%%%%%%%%%%%%%%%%%%%%%%%%%%
%\newpage
%\appendix
%\onecolumn
%\section{You \emph{can} have an appendix here.}
%
%You can have as much text here as you want. The main body must be at most $8$ pages long.
%For the final version, one more page can be added.
%If you want, you can use an appendix like this one.  
%
%The $\mathtt{\backslash onecolumn}$ command above can be kept in place if you prefer a one-column appendix, or can be removed if you prefer a two-column appendix.  Apart from this possible change, the style (font size, spacing, margins, page numbering, etc.) should be kept the same as the main body.
%%%%%%%%%%%%%%%%%%%%%%%%%%%%%%%%%%%%%%%%%%%%%%%%%%%%%%%%%%%%%%%%%%%%%%%%%%%%%%%
%%%%%%%%%%%%%%%%%%%%%%%%%%%%%%%%%%%%%%%%%%%%%%%%%%%%%%%%%%%%%%%%%%%%%%%%%%%%%%%

\end{document}